\documentclass[letterpaper, 10pt, conference]{ieeeconf}      % Use this line for a4 paper
\usepackage{multirow}
\usepackage{cite}
\usepackage{amsmath,amssymb,amsfonts}
\usepackage{textcomp}
\usepackage{times}  %Required
\usepackage{helvet}  %Required
\usepackage{courier}  %Required
\usepackage{url}  %Required
\usepackage{graphicx}  %Required
\usepackage{multirow}
\usepackage{multicol}
\usepackage{bm}
\usepackage{booktabs}
\usepackage{amsmath}
\usepackage{amssymb}
\usepackage{mathrsfs}
\usepackage{algorithm}
\usepackage{algorithmic}
\usepackage{color}
\usepackage{comment}
\usepackage{tabularx,threeparttable}
\usepackage[colorlinks,
citecolor=blue,
linkcolor=blue,
urlcolor=blue
]{hyperref}

\IEEEoverridecommandlockouts                              % This command is only needed if 
\title{\LARGE \bf
Domain Adaptive Robotic Gesture Recognition with \\ Unsupervised Kinematic-Visual Data Alignment
}

\author{Xueying Shi, Yueming Jin, Qi Dou, Jing Qin, and Pheng-Ann Heng %<-this % stops a space
\thanks{X. Shi, Y. Jin, Q. Dou and P. A. Heng are with the Department of Computer Science and Engineering, The Chinese University of Hong Kong. Q. Dou and P. A. Heng are also with the T Stone Robotics Institute, The Chinese University of Hong Kong. \textit{Corresponding author at: Yueming Jin (ymjin@cse.cuhk.edu.hk)}}
\thanks{J. Qin is with Centre for Smart Health, School of Nursing, The Hong Kong Polytechnic University.}
}

\begin{document}

\maketitle
\thispagestyle{empty}
\pagestyle{empty}

\begin{abstract}
	Automated surgical gesture recognition is of great importance in robot-assisted minimally invasive surgery. However, existing methods assume that training and testing data are from the same domain, which suffers from severe performance degradation when a domain gap exists, such as the simulator and real robot. 
	In this paper, we propose a novel unsupervised domain adaptation framework which can simultaneously transfer multi-modality knowledge, i.e., both kinematic and visual data, from simulator to real robot.
	It remedies the domain gap with enhanced transferable features by using temporal cues in videos, and inherent correlations in multi-modal towards recognizing gesture. 
	Specifically, we first propose an MDO-K to align kinematics, which exploits temporal continuity to transfer motion directions with smaller gap rather than position values, relieving the adaptation burden.
	Moreover, we propose a KV-Relation-ATT to transfer the co-occurrence signals of kinematics and vision. Such features attended by correlation similarity are more informative for enhancing domain-invariance of the model.
	Two feature alignment strategies benefit the model mutually during the end-to-end learning process.
	We extensively evaluate our method for gesture recognition using DESK dataset with peg transfer procedure.
	Results show that our approach recovers the performance with great improvement gains, up to 12.91\% in ACC and 20.16\% in F1score without using any annotations in real robot.

\end{abstract}
\section{INTRODUCTION}
Robot-assisted Minimally Invasive Surgery (RMIS) increases the precision and reliability of surgical manipulation with improved control.
Developing autonomy in RMIS can help to facilitate the surgery efficiency and ergonomic comfort of surgeons and patients~\cite{yang2017medical}.
Automated surgical gesture recognition, aiming at automatically perceiving the current action units within surgical tasks, serves as a key prerequisite for surgical automation~\cite{murali2015learning,preda2016cognitive,reiley2010motion,nagy2019dvrk}.
Additionally, the contextual awareness of the surgical state has various applications in post-operative analysis, such as automatic skill evaluation~\cite{reiley2009task}, and surgery educational training~\cite{ahmidi2017dataset}.
%, such as peg transfer, suture, and needle passing\cite{hung2013comparative}. 
%So, understanding and predicting the next surgical actions is essential for autonomous surgery. 

Machine learning based approaches have achieved great successes on surgical gesture recognition task~\cite{ahmidi2017dataset,jin2017sv,van2020multi,gao2020automatic}.
These existing approaches are mostly under the assumption that both training and testing data are from the same distribution, which is generally collected from the real robot.
However, plenty of recent studies have pointed out the problem of performance degradation when encountering domain shift.
In the context of the surgical robot with multiple resources of vision and kinematic, the situation of heterogeneous domain shift is even more severe, given the different robotic configurations, diverse workspaces for surgeon training, and surgeon operative skills.
To recover model performance, an easy way is to re-train models with large-scale labeled data from each domain.
However, collecting and annotating data for every new domain is obviously and prohibitively expensive.

\begin{figure}[t!]
	\centering
	{\includegraphics[width=1\linewidth]{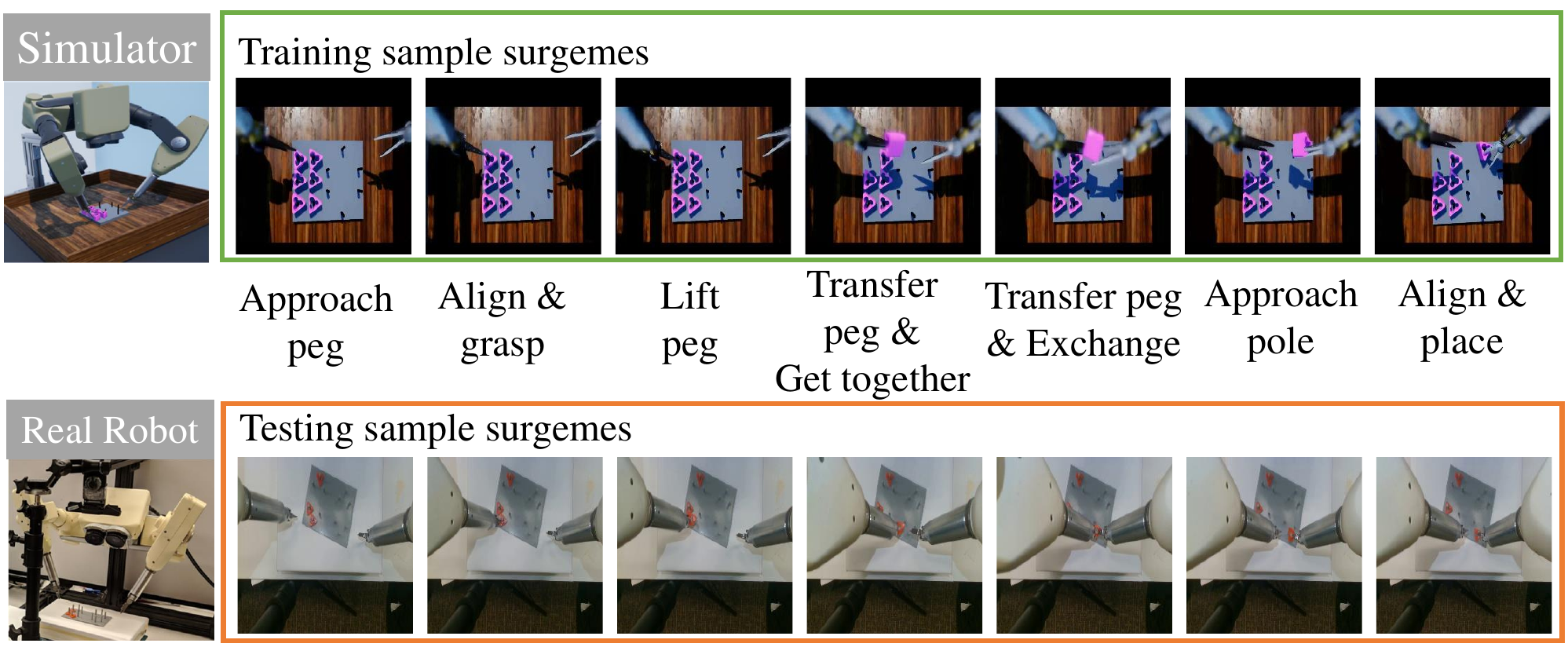}}
	\vspace{-6mm}
	\caption{Seven gestures (surgemes) in peg transfer task with respect to simulator and real-robot environments, where two scenarios are with large domain shift of kinematic and visual information.}
	\label{Fig:TarusVSSim}
\end{figure}

Robotic simulator provides a relatively easy fashion for data collection in a scalable manner, given its economical and safe environment with unlimited maneuver time~\cite{tobin2017domain}.
Besides, The simulator needs no additional annotation to recognition activities of the video because the label is natural generated simultaneously with the given gesture command.
%Simulated surgical environment provides a relatively easy fashion for data collection given the following reasons.
%First, simulator is economical and safe environment, which allows researchers to collect enormous amount of data at a lower cost and in a scalable manner~\cite{tobin2017domain}.
%Second, it provides unlimited amount of training time unlike the expensive single-use mock models~\cite{peng2018sim}.
%Third, 
Simulated environment also alleviates the data privacy concern in the robotic-assisted surgery performed on patients.
%Moreover, the simulator manipulating robotic tools simultaneously comes with the gesture ground truth, without the further requirement of manual annotations. 
Transferring the knowledge with unsupervised domain adaptation (UDA) from simulator to real-robot (Sim2Real) can bring great benefits for tackling label efficiency in real-world RMIS, as annotations from real-robot scenario are not the requisite.
However, effectively transferring the Sim2Real knowledge of surgical gesture is quite challenging given extreme visual gap between the simulator and real-robot as shown in Fig.~\ref{Fig:TarusVSSim}.
In robotic peg transfer task, a task holding primary importance in surgical skill learning, unidentical peg numbers and tilt degree of pegboard further increase the visual domain gap.
%given the extreme visual gap between the simulator and real-robot as shown in Fig.~\ref{Fig:TarusVSSim}.

%yosinski2014transferable,li2016revisiting,%tzeng2014deep,taigman2016unsupervised,
%Approaches include re-training the model in the target domain~\cite{yosinski2014transferable}, adapting the weights of the model based on the statistics of the source and target domains~\cite{li2016revisiting}, learning invariant features between
%domains~\cite{tzeng2014deep}, and learning a mapping from the target domain to the source domain~\cite{taigman2016unsupervised}.
The computer vision community has devoted significant study to the problem of adapting vision-based models trained in a source domain to a previously unseen target domain~\cite{li2018adaptive,
	saito2018maximum
}, which is highly related to domain adaptation from simulator to real robot in robotic perception tasks.
Recently, tackling the visual gap from simulator to real-robot attracts the increasing interests in the robotic-related tasks.
For example, 
%several methods for data-efficient robotic grasping propose to input a single image frame to predict the next state of robot and transfer the learned policy from simulator to real robot~\cite{tobin2017domain,james2017transferring,bousmalis2018using,james2019sim,hundt2020good, jeong2020self,iqbal2020toward}.
%recent methods for data-efficient robotic grasping propose to tackle the visual gap from simulator to real-robot (Sim2Real) in image-level~\cite{james2017transferring,james2019sim,hundt2020good}.
%These methods input a single image frame to predict the next state of robot and transfer the learned policy from simulator to real robot.
%In addition, 
domain adaptation methods with pixel-to-pixel transformation from simulated surgical scene to real one are proposed for surgical instrument segmentation task~\cite{sahu2020endo}.
Although achieving the promising performance, purely aligning image-level features fails to well handle the adaptation of dynamic gestures in robotic videos.
% bridge the dynamic 
Unsupervised domain adaptation for action recognition in natural videos is an emerging research topic.
Recent studies point out that narrowing the feature discrepancy in both spacial and temporal spaces can outperform aligning features in single spacial space~\cite{chen2019temporal,munro2020multi,choi2020shuffle,pan2020adversarial}.
%For example, \cite{chen2019temporal} xxxxxx
These successes inspire us to extract the domain-invariant spatial-temporal features for Sim2Real robotic gesture recognition.
However, visual gap between simulator and real robot is extreme larger than natural computer vision tasks, so only relying on transferring the visual modality is not enough for good domain adaption.

Apart from vision information, the kinematic data, a unique resource recorded from robot system, plays the key role for robotic surgical gesture recognition.
Several works propose to develop multi-modal learning methods to leverage the complementary cues contained in the video vision data and kinematics for accurate gesture recognition~\cite{lea2016learning,qin2020temporal}.
However, in the context of domain adaptation, there also exists a large domain shift of kinematic information between simulator and real-robot.
For example, the relative distances between robot arm and pegboard, also, between pegs in peg transfer task show differences, leading to different trajectory lengths on two platforms for the same gesture.
How to leverage the inherent correlation to bridge multi-modal domain shift in robotic videos is essential for Sim2Real gesture recognition, yet has not been explored so far.
\begin{comment}
In this paper, we propose a novel multi-modal domain adaptation framework to transfer knowledge from simulator to real-robot for surgical gesture recognition.

\end{comment}

\begin{figure*}[ht!]
	\centering
	{\includegraphics[width=1\linewidth]{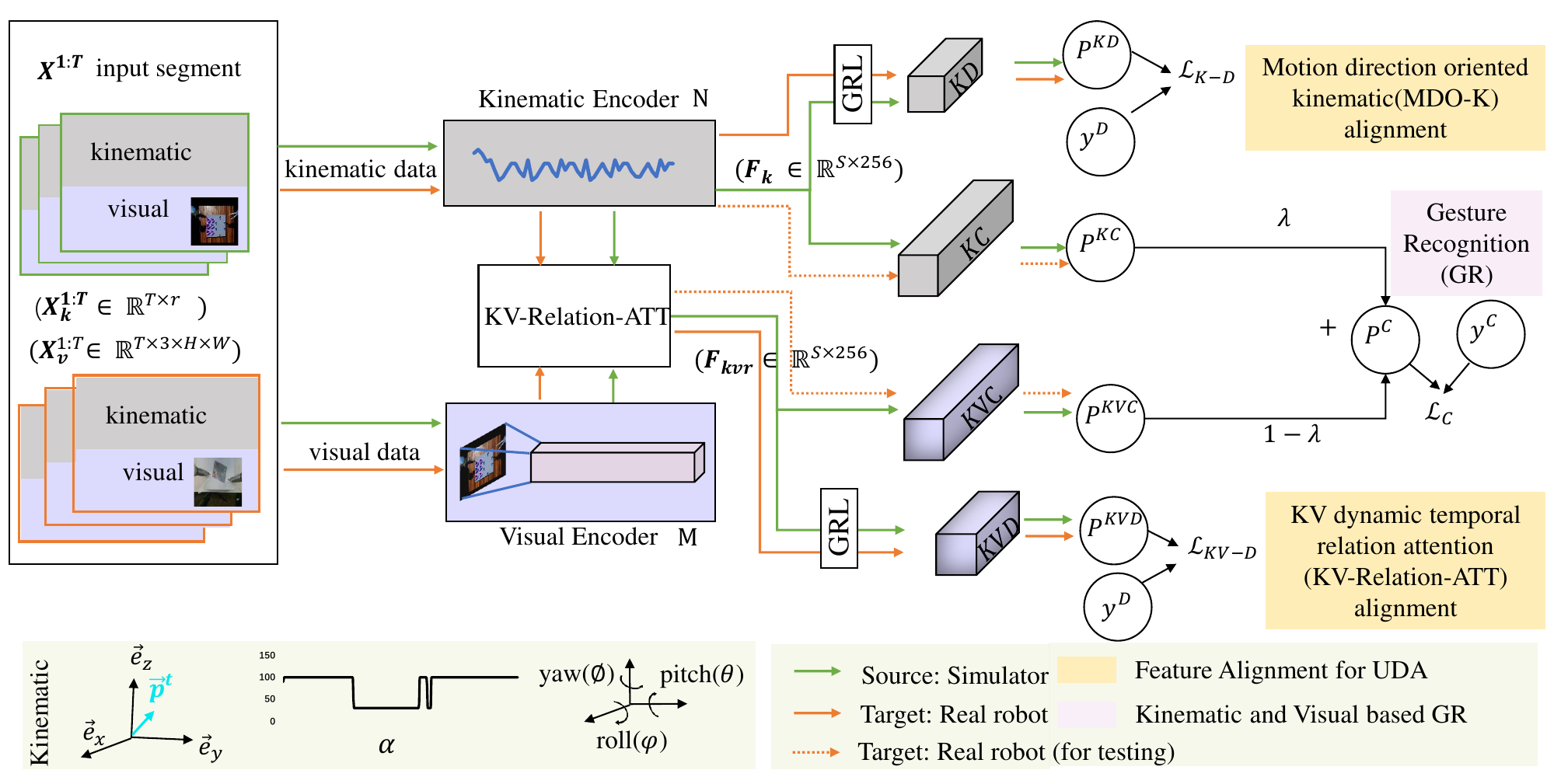}}
	\caption{The proposed KV-Sim2Real framework for unsupervised domain adaptation from simulator to real robot in peg transfer gesture recognition task. The network conducts three tasks: two feature alignment (FA) tasks for UDA (yellow box) and one multi-modality based gesture recognition (GR) task (pink box).  $X_k^{1:T} \in \mathbb{R}^{T\times r}$ is the kinematic input segment($r$ for dimension of each frame, T for segment length) while $X_v^{1:T} \in \mathbb{R}^{T\times3\times H \times W}$ is the corresponding visual segment($H$ and $W$ are height and width of RGB image). We denote the source (green arrow) and target (orange arrow) domain as simulator and real robot respectively, while they share the same kinematic encoder $N$, the Visual Encoder $M$ and KV-Relation-ATT module. Gradient Reversal Layer (GRL) is designed for domain alignment with adversary training and utilized in two domain discriminators KD and KVD. KD and KVD and designed based on the input representations, with KD representing the Kinematic feature based Discriminator and KVD representing the Kinematic and Visual fusion feature based Discriminator. KC and KVC are classifiers for multi-modality gesture recognition. $p$ and $y$ are prediction probability and label. $\mathcal{L}_{K-D}$ and $\mathcal{L}_{KV-D}$ are losses for KD and KVD while $\mathcal{L}_{C}$ is loss for GR.}
	\label{Fig:pipeline}
\end{figure*}
In this paper, we aim to handle above domain gap which can come from
variable environments (simulator and real robot) and minor shifts in robot hardware during identifying surgemes performed by a surgeon through teleoperation. The main contributions are as follows:

\begin{itemize}
	\item We present a novel unsupervised domain adaptation framework, which can simultaneously transfer multi modality knowledge (kinematic data as well as visual data) from simulator to real robot to remedy the domain gap for surgical gesture recognition. 
	%The whole process is conducted in unsupervised domain adaptation manner where gesture labels are only available at simulator domain.
	
	%We introduce a domain adaptation framework which can synchronized transfer multi modality knowledge (kinematic data as well as visual data) from simulator to real robot in order to remedy the domain gap between them for peg transfer task on DESK dataset. The whole process is conducted in unsupervised domain adaptation manner where surgeme recognition labels are only available at simulator domain.
	
	\item We propose the motion direction oriented strategy to transfer the kinematic information (MDO-K).
	Bypassing the network overfitting in specific position values, the variables of direction vector are utilized to encourage the network to extract enhanced domain-invariant representations by incorporating the temporal cues.
	
	\item We introduce a KV-Relation-ATT embedding which can measure the co-occurrence between the kinematic data and visual data at each time step. 
	Then we transfer this KV-Relation-ATT to make complementary of visual information and kinematic at each time step. 
	
	\item To the best of our knowledge, we are the first to explore how to decrease both kinematic and visual gap between simulator and real robot. Our framework can outperform the state-of-the-art UDA algorithms and decrease the domain gap by 12.91\% in ACC compared with trained only on simulator and test on real world.
	
\end{itemize}

\section{METHODS}
\subsection{Problem Definition and Data Setup}
%%DA task, unsupervised, task + DA training manner
Our method aims to perform transfer learning in gesture recognition from simulator to real robot, which stands on an unsupervised domain adaptation setting, given no gesture label in the real-robot environment, instead, the gesture recognition would only be supervised by simulator label during training. 
The network is trained in a multi-task manner doing gesture recognition (GR) as well as feature alignment (FA) at the same with the aim of embedding the feature as a domain-irrelevant feature for gesture recognition.

The dataset we utilized is Dexterous Surgical Skills Transfer (DESK) dataset~\cite{madapana2019desk}, which contains peg transfer motions recorded from a simulated Taurus II robot, and a real Taurus II robot.
The motion generally consists of picking an object from a pegboard with one robotic arm, transferring it to the other arm, and positioning the object over a target peg on the opposite side of the board. 
The definition of detailed seven gestures is shown in Fig.~\ref{Fig:TarusVSSim}.
We employ the RGB recorded video and kinematic information of robotic arms to form the multi-modality input resource.
Each training or testing sample is defined by a video `segment', which consists of a variable number of continuous frames holding the same gesture label. %in peg transfer gesture recognition task
Each item in the video segment comprises a video frame and corresponding kinematic information. 

The kinematic data consists of 14 kinematic variables per frame that represent the robot’s end-effector pose(7 features in each arm). It includes the wrist orientation (yaw, pitch and roll
angles), the translation in x, y and z coordinates with respect to the robot’s origin and the griper state (also a value between 30 and 100) for both the arms.

\subsection{KV-Sim2Real Gesture Recognition UDA Framework}
% The multi-task setting, the input of training and testing, the alignment
An overview of our proposed method for multi-modal domain adaptation in robotic surgical gesture recognition is shown in Fig.~\ref{Fig:pipeline}.
%Fig.~\ref{Fig:pipeline} is the Sim2Real gesture recognition UDA pipeline conducting on multi-modality: kinematic and visual. 
In the training process, the network conducts three tasks: two feature alignment (FA) tasks for UDA (yellow box) and one supervised multi-modality based gesture recognition (GR) task (pink box) given the simulator gesture label. 
After finishing training, when conduct gesture recognition on real robot, the trained model has generalization ability to recognize the gesture, and be unaffected on training with only simulator gesture label but without real robot label.

More specifically, we denote data from the source (green arrow) and target (orange arrow) domain as $I_s$ and $I_r$ for simulator and real robot respectively. 
The segment inputs $X^{1:T} = (X_k^{1:T}, X_v^{1:T}) \in {I_s \cup I_r}$ share the visual encoder $M$ and kinematic encoder $N$ to extract features. 
The kinematic and visual features are trained to be domain-irrelevant by FA, which is achieved by the adversarial training with the discriminator (KD / KVD) to differentiate which domain the input comes from. The two types of discriminators are designed with respect to its input representations, with KD representing the Kinematic feature based Discriminator and KVD representing the Kinematic and Visual fusion feature based Discriminator.

%(KD / KVD, KD is kinematic feature based discriminator while KVD is kinematic and visual relation feature based discriminator, details would be introduced in later sections) that predicts the domain. 
%A Gradient Reversal layer (GRL) reverses and back-propagates the gradient to the features. 

We align both kinematic and visual features which are all with domain shift between the simulator and real robot. In the next several sections, we introduce our method for feature alignment, as well as how to conduct gesture recognition with feature alignment together. 
We first propose Motion Direction Oriented Kinematics (MDO-K) feature alignment which aligns kinematics using features of motion direction.
Second, we propose a KV dynamic temporal relation attention (KV-Relation-ATT, KV stands for kin and vis) alignment to highlight the co-occurrence KV feature by picking out the visual data whose feature embeddings have a high degree of synchronization with the kinematic feature.

\begin{comment}
\begin{table}[htbp]
\centering
\caption{Kinematic variables in simulator and real robot we utilize. Note that \textit{ts} is the Unix timestamp,
$\vec{J}$ is the vector of joint angles, $\vec{p}$ is the position vector (x, y
and z),  $\vec{\theta}$ be the Euler angles (yaw, pitch and roll), \textit{gs} is the gripper
state of the end-effector and \textit{R} be the $3 \times 3$ rotation matrix. }
\begin{tabular}{c|c||c|c}
\hline
\multicolumn{2}{c||}{\textbf{Taurus II}} & \multicolumn{2}{c}{\textbf{Taurus II Simulator}}  \\
\hline
\hline
ID         &Variable     &ID         &Variable \\
\hline
\hline
1          &\textit{ts}           &1          &\textit{ts} \\
2-13       &\textit{R} and $\vec{p}$      &2-4        &$\vec{p}$ \\
& -           &5-7        &$\vec{\theta}$ \\
14-16      &$\vec{p}$            &8-14       &$\vec{J}$  \\
17         &\textit{gs}           &15         &\textit{gs} \\
\hline
\end{tabular}

\label{table:1}
\end{table}
\end{comment}

\begin{comment}
\begin{figure}[ht]
\centering
{\includegraphics[width=1\linewidth]{figures/KV_input.pdf}}
\caption{The kinematic and visual input information. }
\label{Fig:KV_input}
\end{figure}
\end{comment}

\subsection{Motion Direction Oriented Kinematics Feature Alignment}
Effectively utilizing temporal information in robotic videos is crucial for enhancing the kinematic feature adaptation.
We propose to align the kinematics features oriented for motion direction, instead of straightforwardly using the specific position vector.

%which is a significant factor of peg transfer. 
We denote the input kinematic data as $X_k^{1:T} \in \mathbb{R}^{T\times r}$, where $T$ denotes the frame number of a sample segment, $r$ is the vector length with the vector consisting of all kinematic information as shown in Fig.~\ref{Fig:pipeline}, including Taurus robot's end-effector position $\vec{p} = (a,b,c)$, end-effector orientation $(\varphi,\phi,\theta)$ representing (yaw, pitch, roll), and grasp angle $\alpha$ for both robotic arms. 
For motion oriented kinematic feature modeling, position vector $\vec{p}$ (here $\vec{p}$ equals to coordinate value under the rectangular coordinates, where the rectangular coordinates are the same for both simulator and real robot) would be better than the end-effector orientation and grasp angle, because the last two parameters can only model the dexterous skill of the end-effector focusing on grasping moment and is not suitable for motion direction modeling of the whole grasp process compared with position vector.

However, the position settings for objects is more complex in real world than that in simulator, such as between-object distance shift (peg from peg/robot arm from pegboard) existing between source domain and target domain, the robotic arms' trajectory would be different even conducting the same gesture between same nearby pegs.
\begin{comment}
Therefore if we directly use the coordinate $\vec{p}$, 
the network cannot learn the motion of the robotic arm, instead, it may focus on learning the exact start and end position value.
So, in real world where the distance between pegs or between robotic arm and pegboard has variable shift, it's hard to recognize the same gesture in real world because the kinematic information is trained to overfit on simulator without the ability to handle such distance variance.  
\end{comment}

\begin{figure}[ht]
	\centering
	{\includegraphics[width=0.56\linewidth]{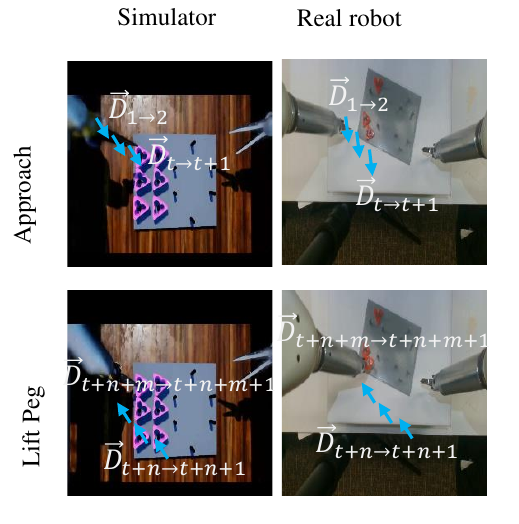}}
	\caption{Gestures are easy to be distinguished between `approach' and `lift peg' using direction vectors in both simulator and real robot since those two gestures are in opposite directions. Therefore aligning the motion direction of the robot arm (represented by unit direction vectors) can help transfer gestures from simulator to real robot. }
	\label{Fig:unit_direction_vector}
\end{figure}

In order to handle between-object distance shift, we propose to replace position vector $\vec{p}$ with relative direction vector $\vec{D}_{t\rightarrow t+1} = \vec{p}_{t+1}-\vec{p}_{t}$, which represents the positional relationship between any two points in space (see Fig.~\ref{Fig:unit_direction_vector}). 
Compared to the absolute position vector $\vec{p}_{sim}$ and $\vec{p}_{real}$, the relative position vector $\vec{D}_{t\rightarrow t+1}$ can model the motion direction of the robotic end-effector independently with which start point and end point they are. 
Therefore, the relative direction vector would avoid the issues of different start and endpoints of different pegs in the peg transfer task. 
Transferring position vector to relative direction vector within kinematic data can greatly narrow the domain gap between the simulator and real robot, helping to encourage the network to extract domain-invariance features by relieving the learning burden. We normalize the vector as:
\begin{equation}
	\begin{aligned}
		\label{equ:1}
		\vec{D}^{unit}_{t\rightarrow t+1} &= \frac{\vec{D}_{t\rightarrow t+1}}{|\vec{D}_{t\rightarrow t+1}|}.
	\end{aligned}
\end{equation}

\begin{comment}
\vec{D}_{unit} &= \frac{a{\vec{e_x}} + b{\vec{e_y}} + c{\vec{e_z}} }{\sqrt{x^2 + y^2 + z^2}} \\
&=(\frac{a}{\sqrt{a^2 + b^2 + c^2}}{\vec{e_x}},\frac{b}{\sqrt{a^2 + b^2 + c^2}}{\vec{e_y}},\frac{c}{\sqrt{a^2 + b^2 + c^2}}{\vec{e_z}})
\end{comment}
The $(\vec{D}_{1\rightarrow 2}^{unit},... \vec{D}_{{T-1}\rightarrow {T}}^{unit})$ can model the continuous between-frame motion directions while ignoring relative distance, forming as a trajectory of the robotic arm and integrating the temporal information from the videos.

Our input kinematics of one single frame are then transformed as $x_k^{t}=(\vec{D}^{unit}_{t\rightarrow t+1},\varphi,\phi,\theta, \alpha)$, with $7$ elements for each side of robotic arms, and in total $14$ elements.
Then we forward $x_k^{t}$ into the kinematic encoder for extracting features $F_k$, followed by a Gradient Reversal Layer (GRL)~\cite{ganin2015unsupervised} unit and kinematic discriminator $KD$ to do domain alignment. The GRL is designed for inverting the gradients during training the domain discriminators such that the learned feature would not overfit the source domain.
The loss for motion direction oriented alignment (MDO-K) is defined as follows:
\begin{equation}
	\mathcal{L}_{K\text{-}D}^{v} = \mathcal{L}_{d}(KD( N(x_{k}^{1:T}) ), y^{D}),
	\label{equ:lossKD}
\end{equation}
where $v$ represents one sample segment, with $v\in {I_s \cup I_r}$ ($I_s$ for simulator, $I_r$ for real robot); $\mathcal{L}_{d}$ is domain discriminator loss; $y_{D}$ is domain label with 0 for the samples from simulator and 1 for the samples from real robot.
\begin{comment}
For KV alignment and kinematic alignment, we aim to learn an embedding $K^{T\times F}$ from $K^{T\times r'}$ by Kinematic Encoder  The input for kinematic type data is $K^{T\times r'}$ with $T\times r'$ dimension, as shown in Fig.~\ref{Fig:pipeline}.
\end{comment}

\begin{comment}
{In order to handle between-object distance shift, we propose to replace position vector $\vec{p} = (a,b,c)$ with unit relative direction vector $\vec{D}_{t_{i}\rightarrow t_{i+1}} = \vec{p}_{t+1}-\vec{p}_{t}$, which represent the positional relationship between any two points in space. Compared to the absolute position vector $\vec{p}_{sim}$ and $\vec{p}_{real}$, the relative position vector $\vec{D}_{t_{i}\rightarrow t_{i+1}}$ can model the motion direction of the robotic end-effector independently with which start point and end point they are. So transfer position vector to relative position vector within kinematic data can shorten the domain gap between the simulator and real robot. 
\end{comment}
\begin{comment}
Most previous works utilize single modality (visual / kinematic) as input for gesture recognition. 
The limitation of single-modality action recognition models is that either kinematic or visual input is representative feature, because they describe a given moment's gesture from single perspective. In order to utilize multi-modality feature, we propose to conduct Kin-Vis (KV) representation.
\end{comment}

\subsection{KV Dynamic Temporal Relation Attention Alignment}
Recent methods have found that combing kinematic and visual (KV) features would promote gesture recognition in a multi-task training manner~\cite{qin2020temporal,rahman2019transferring}.
The reasons are as follows: kinematic and visual information is with a complementary relationship and would adding supplementary information for each other when one modality can not work independently. The kinematic has $3\text{D}$ information such as position vector and rotation of the robotic arm's end-effector, which can recognize existing state by `manipulating'. Meanwhile, the visual information can provide $2\text{D}$ image-based environment setting, such as the relative position between moving robotic arm and pegboard, which can recognize existing state by `seeing'. For a situation like occlusion between robotic arm and the peg during the peg transfer process, the extracted kinematic parameter at this moment would not be interrupted by visual occlusion and can remedy this problem; For a situation like the robotic arm is moving back and forth for a relatively long time because some trainers are with not enough proficiency during manipulation of the robotic arms, leading to the parameters extracted from robotic arm be unrepresentative. The visual relative position information between objects (robotic arm and peg/peg and pegboard) can remedy this situation and would tell what gesture is it now (such as success grasp or not), which is more accurate compared with random and irregular kinematic motion data at this moment. 

However, KV feature cannot work in DA setting if kinematic and visual features are aligned separately in a multi-task manner. The reason comes from that the visual domain gap is very large between simulator and real robot compared to kinematic domain gap, separately align kinematic and visual features would lead to model overfit to the visual source domain, so the performance would decrease when conducting multi-modality gesture prediction in real robot. In order to handle this, we propose a new domain align pattern oriented for multi-modality setting: KV-Relation attention alignment (KV-Relation-ATT). The motivation is to propose a more strict constraint feature for multi-modality domain alignment: co-occurrence feature alignment. We aim to align the co-occurrence feature of kinematic and visual signal instead of any single signal occurrence. Next, we would introduce the KV embedding and how to use it for KV-Relation-ATT DA computation.
\begin{comment}
It is worth mentioning that, unlike methods who conduct kinematic and visual feature representation separately and use multi-task learning manner to combine these two modality features~\cite{qin2020temporal,rahman2019transferring}, we combine these two modality feature by consider the relation similarity between kinematic data as well as visual data at the same moment within a video segment $X^{1:T}$. As shown in Fig.~\ref{Fig:KVcompute},
\end{comment}

\begin{figure}[htbp]
	\centering
	{\includegraphics[width=1\linewidth]{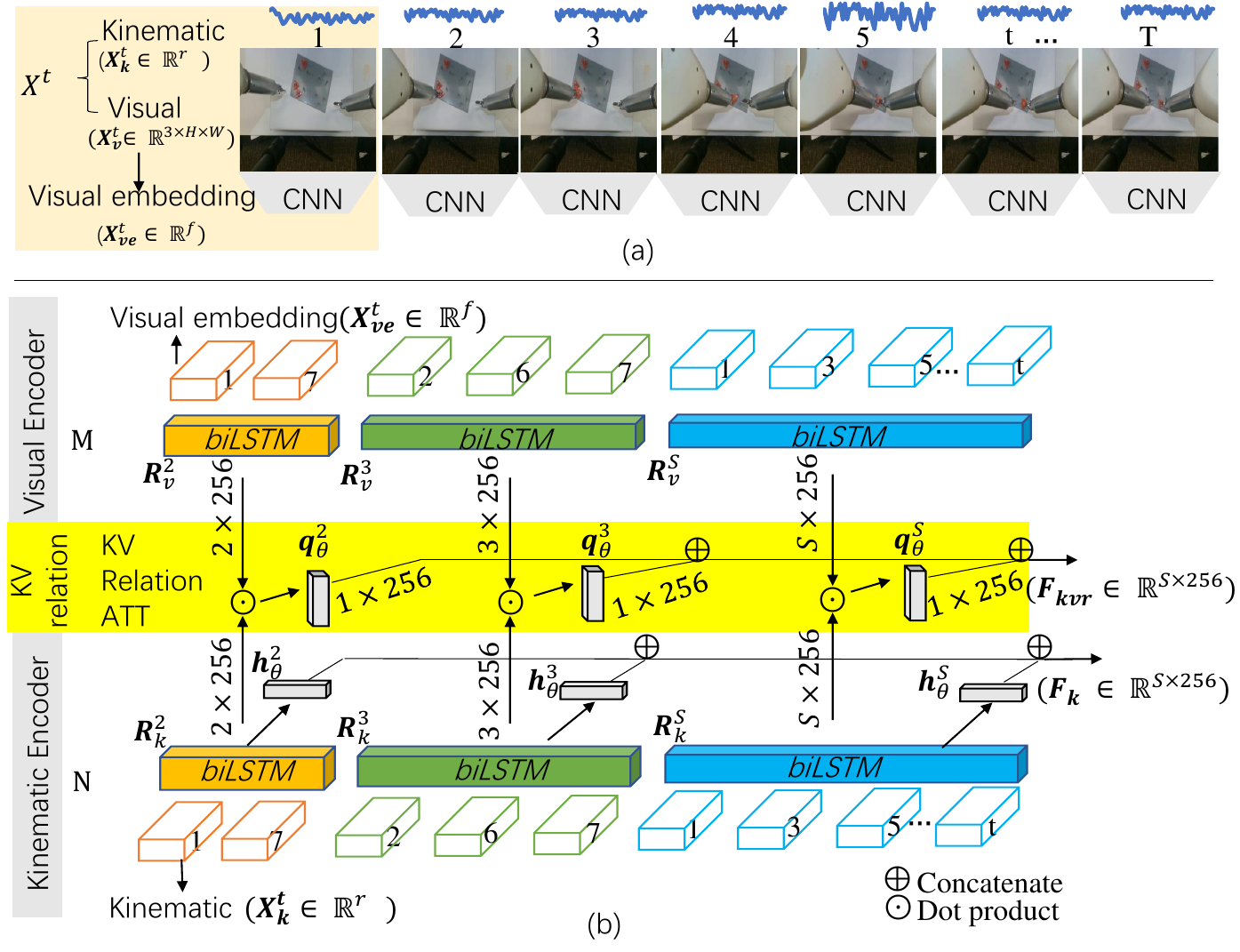}}
	\caption{KV Dynamic Temporal Relation Attention (KV-Relation-ATT), the co-occurrence of the kinematic and visual information is interpreted by the dot-product of each modality feature at each time $t$.}
	\label{Fig:KVcompute}
\end{figure}

The KV embedding is as follows. The input is video segment $X^{1:T}$, as shown in Fig.~\ref{Fig:KVcompute}. For a given input segments $X^{1:T} = (X_k^{1:T},X_v^{1:T})$ in Fig.~\ref{Fig:KVcompute}\textcolor{blue}{(a)}, $X_k^{1:T}$ is a kinematic input segment and $X_{ve}^{1:T}$ is its corresponding visual input segment. Each visual input data $X_{ve}^{t}$ is the feature extracted by pre-trained ResNet101 model~\cite{he2016deep} from original input data $X_{v}^{t}$. Here, in order to get the high level semantic visual feature, we use the bottom feature layer before the last fc layer of ResNet101 with dimension $f$($X_{ve}^{t}\in \mathbb{R}^{f}$).  For kinematic and visual feature embedding, we utilize temporal-wise embedding with the reason that temporal representation outperforms frame-wise ones for action recognition, so we final utilize the TRN (Temporal Relation Net)~\cite{Zhou_2018_ECCV} for both kinematic and video temporal feature embedding, represented by visual encoder $M$ and kinematic encoder $N$ with input $X_{ve}^{1:T}$ and $X_k^{1:T}$, as shown in Fig.~\ref{Fig:KVcompute}\textcolor{blue}{(b)}.
Activity in video sequence is highly related to temporal relations, which is the meaningful object variation along times. So, how to define the temporal relation would be significant for video sequence feature representation. As the temporal relation would exist in short-term or long-term variations. In this regard, we capture temporal features at multiple time scales, where `scale' corresponds to how many frame numbers we sample from a given sequence. the TRN would embed a given segment $X^{1:T}$ with temporal relation of different time scales $S$ by biLSTM ($R_v^s$ for visual, $R_k^s$ for kinematic).
\begin{comment}
In addition, in order to handle domain variance during our peg transfer task, we propose to align relation attention between kinematic and visual data with dynamic temporal. If we separately aligning kinematic feature $R_k^n$ or visual feature $R_v^n$, it can not extend the multi modality signal to the largest extend. Which means, the tranfer upper bound performance is the adding of their single tranfer performance. So, we propose to focus on the co-occurring feature between kinematic and visual in a same moment, which means the object motion can be interpreted by kin-vis temporally synchronized degree. 
\end{comment}
\begin{comment}
\begin{equation}
\label{equ:2}
T_v(V) = h_{\theta}(\sum_{i<j}^{}g_{\theta}(f_i,f_j)),

\end{equation}
\end{comment}

The implement of the KV-Relation-ATT alignment is as follows: we set input video segment $X^{1:T}$ with different scales of ordered frames as $x_{ve}^{1},...x_{ve}^{T}$, where $x_{ve}^{t}$ is a representation of the $t^{th}$ frame of the segment. The co-occurrence of the kinematic and visual information is interpreted by the dot-product of each modality feature at each time $t$. Then, the two-frame-scale KV-Relation-ATT is interpreted as attention format of kinematic and visual feature:
\begin{equation}
	\label{equ:3}
	F_{kvr}^2 = R_v^2 (x_{ve}^{i},x_{ve}^{j}) \cdot  R_k^2 (x_{k}^{i},x_{k}^{j}), (i<j)
\end{equation}
where $i$ and $j$ are time indexes sampled from video segments.

The multiple time scale KV-Relation-ATT feature format is to use the following composite function to accumulate frame relations at different scales:
\begin{equation}
	F_{kvr} = q_{\theta}^2(F_{kvr}^2) + q_{\theta}^3(F_{kvr}^3)...+ q_{\theta}^S(F_{kvr}^S),
\end{equation}
where $q_{\theta}^S$ is a fully connected layer which transforms different scale feature as the same dimension in order to concatenate them together as a multiple time scales KV-Relation-ATT feature $F_{kvr}$.
\begin{comment}
Each relation term $T_d$ captures temporal relationships between $d$ ordered frames. Each $T_d$ has its own separate $h_{\phi}^(d)$ and $g_{\theta}^{(d)}$. Notice that for any given sample of $d$ frames for each $T_d$, all the temporal relation functions are end-to-end differentiable, so they can all be trained together with the base CNN used to extract features for each video frame.
\end{comment}

For visual information, our pre-trained CNN model (here we utilize ResNet101 as network backbone) first embed visual input frame $x_v^{t}$ at time $t$ into visual feature $x_{ve}^{t}$ and fixed it without further fine-tuning. For kinematic information, with the reason of lacking of pre-trained model for kinematic, we direct use $x_k^{t}=(\vec{D}^{unit}_{t\rightarrow t+1},\varphi,\phi,\theta, \alpha)$ as input define in the end of Sec.~C. 
\begin{comment}
The kinematic-visual dynamic temporal relation attention alignment is to distinguish whether the feature $F_{kvr}$ comes from the simulator or real robots. We combine a gradient reversal layer (GRL) with a binary domain discriminator to form $KD$.  Through the adversarial objectives, the discriminators are optimized to classify different domains, while the feature extractors are optimized in the opposite direction. 
\end{comment}

The loss for DA is as follows ($KV$ represent KV-Relation-ATT calculation, $u$ is one sample segment, $u\in {I_s \cup I_r}$, $y^{D}$ is domain label (0 / 1) ):
\begin{equation}
	\mathcal{L}_{KV\text{-}D}^{u} = \mathcal{L}_{d}(KVD( KV (M(x_{ve}^{1:T}), N(x_{k}^{1:T})) ), y^{D})
	\label{equ:lossKVD}
\end{equation}

\subsection{Overall Loss Function and Training Details}
During the training of gesture recognition task, we do gesture recognition along with domain adaption (two alignment methods introduced in C. and D. ). We input source domain data (simulator) and target domain data (real robot) simultaneously, each domain has the same batch size. Only source domain would calculate the gesture recognition classification loss which is as follows ($w$ is one sample segment, $w\in {I_s}$, $\mathcal{L}_{ce}$ is cross-entropy loss, $y^c$ is gesture label, $p^{kc}$ and $p^{kvc}$ are probabilities generated from network):

\begin{equation}
	\begin{array}{ll}
		\mathcal{L}_{C}^{w} &= \mathcal{L}_{ce}(p^c,y^c), \\
		p^c &= \lambda p^{kc} + (1-\lambda) p^{kvc} 
	\end{array}
	\label{equ:CE}
\end{equation}
So the total loss for training is:

\begin{equation}
	\mathcal{L} = \sum_{w\in {I_s}}{\mathcal{L}_C^w} + \sum_{u\in {I_s \cup I_r}}{\mathcal{L}_{KV\text{-}D}^{u}} + \sum_{v\in {I_s \cup I_r}}{\mathcal{L}_{K\text{-}D}^{v}}
	\label{equ:lossTotal}
\end{equation}

The implementation details are as follows. 
We assign the output feature dimension of biLSTM in both the visual encoder and the kinematic encoder as $256$. 
The relation scale in $S$ is set as $10$.
%which consider relation from $2$ frame to $S-1$ frame relation.
%
The batch size for source and target data is $256$ in training, so the total batch number is $512$; During validation, batch size is also set as $512$ with source and target data randomly mixed. 
The reverse decay rate for network back propagation in GRL is set to $0.5$.
We use Adam optimizer to train the model. 
The learning rate is initialized by $1\times 10^{-3}$. 
Our model is implemented based on the PyTorch using one NVIDIA
Titan Xp GPU for acceleration. To avoid the case
of coincidence, we trained the same model for three times and reported their average results.

The input of our KV-Sim2Real architecture is a video segment $X^{1:T} = (X_v^{1:T},X_k^{1:T})$, which consists of variant length frames. 
Every single frame's visual information $X_v^{t}$ is extracted from the video with $1920\times 1080$ resolution ratio.
Each video is captured at $30$ fps, and after cutting the video into video frames, we resize each frame to $224\times 224$. 
We utilize pre-trained Res101~\cite{he2016deep} to directly extract features for visual frame $X_v^{t}$ and take the last embedding layer with size $2048$ as fixed visual embedding without further training. 
All the frames are labeled with 7 defined gestures (shown in Fig.~\ref{Fig:TarusVSSim}).

\section{EXPERIMENTS}
\begin{comment}
\begin{table}[htbp]
\centering
\caption{Overall comparison with the state-of-the-art unsupervised domain adaptation methods (mean$\pm$std., \%). '+' means multi-task learning results. }
\begin{tabular}{l|l|l|l}
& Method    &  ACC         & Gain  \\ \hline
& Baseline    & 46.85\pm1.32  & -     \\ \hline
\multicolumn{1}{c|}{\multirow{2}{*}{\begin{tabular}[c]{@{}c@{}}Image based \\ DA \end{tabular}}} & AdaBN~\cite{li2018adaptive}                                    & xx          & xx    \\ \cline{2-4} 
\multicolumn{1}{c|}{}                                                                                  & MCD~\cite{saito2018maximum}                              & xx          & xx    \\ \hline
\multicolumn{1}{c|}{\begin{tabular}[c]{@{}c@{}}Video based \\ DA\end{tabular}}    & T3AN~\cite{chen2019temporal}                             & 15.79\pm1.36          &--       \\ \hline
\multicolumn{1}{c|}{\begin{tabular}[c]{@{}c@{}}Kin based\\ DA (ours) \end{tabular}}    & MDO-K                              & 55.75\pm0.76          &8.90       \\  
\hline

\multicolumn{1}{c|}{\begin{tabular}[c]{@{}c@{}}KV based \\ DA (ours)\end{tabular}}
& \begin{tabular}[c]{@{}l@{}}MDO-K  \\ +\\ KV-Relation\end{tabular} &57.89\pm0.39             &11.04       \\  \hline

\end{tabular}
\label{table:overallresults}
\end{table}
\end{comment}

\begin{table*}[]
	\centering
	\caption{Overall comparison with the state-of-the-art unsupervised domain adaptation methods (mean$\pm$std., \%). '+' means multi-task learning results. }
	\resizebox{1\textwidth}{!}
	{
		\begin{tabular}{l||l||c|c|c|c|c|c|c|c|c|c}
			\hline
			& Method    &  Accuracy  & Gain &  Precision & Gain &  Recall & Gain &  Jaccard & Gain &F1score & Gain \\ \hline
			
			\multicolumn{1}{l||}{\multirow{2}{*}{\begin{tabular}[c]{@{}c@{}}Without DA \end{tabular}}} & RF(Random Forest)~\cite{rahman2019transferring}                                    & 27.92          & --    &  -- & -- &  -- & -- &  -- & -- &-- & -- \\ \cline{2-12}   
			\multicolumn{1}{l||}{}
			& 
			Baseline(Deep Learning)    & $44.98\pm2.00$  & --   &  $49.09\pm3.89$ & -- &  $44.71\pm2.00$ & -- &  $26.31\pm2.25$ & -- &$37.68\pm2.63$ & --  \\ \hline
			
			\multicolumn{1}{l||}{\multirow{2}{*}{\begin{tabular}[c]{@{}c@{}}Image based \\ DA \end{tabular}}} & AdaBN~\cite{li2018adaptive}                                    & $14.81\pm0.10$          & --    &  $16.14\pm1.83$ & -- &  $14.76\pm0.14$ & -- &  $6.42\pm0.11$ & -- &$11.83\pm0.16$ & -- \\ \cline{2-12} 
			\multicolumn{1}{l||}{}                                                                                  & MCD~\cite{saito2018maximum}                              & $16.03\pm0.83$          & --   &  $15.18\pm2.30$ & -- &  $15.93\pm0.88$ & -- &  $6.26\pm0.33$ & -- &$11.19\pm0.44$ & --  \\ \hline
			\multicolumn{1}{c||}{\begin{tabular}[l]{@{}c@{}}Video based \\ DA\end{tabular}}    & T3AN~\cite{chen2019temporal}                             & $18.23\pm1.08$
			&--      &  $9.95\pm0.83$ & -- &  $18.09\pm1.09$ & -- &  $5.80\pm0.19$ & -- &$10.04\pm0.36$ & --  \\ \hline
			\multicolumn{1}{c||}{\begin{tabular}[l]{@{}c@{}}Kin based\\ DA (ours) \end{tabular}}    & MDO-K                              & $51.17\pm0.10$          &6.19      &  $54.41\pm0.62$ & 5.32 &  $51.27\pm0.06$ & 6.56 &  $35.07\pm0.11$ & 8.76 &$51.11\pm0.07$ & 13.43  \\  
			\hline
			\multicolumn{1}{c||}{\begin{tabular}[l]{@{}c@{}}KV based \\ DA (ours)\end{tabular}}
			& \begin{tabular}[c]{@{}l@{}}MDO-K  \\ +\\ KV-Relation-ATT\end{tabular} &$57.89\pm0.39$             &12.91      &  $60.41\pm0.27$ & 11.32 &  $58.03\pm0.39$ & 13.32 &  $41.58\pm0.41$ & 15.27 &$57.84\pm0.60$ & 20.16  \\  \hline
			
		\end{tabular}
	}
	\label{table:overallresults}
\end{table*}
\subsection{Experimental Setting}
%\textbf{Gesture recognition} We conduct experiments of peg transfer gesture recognition task, where the experimental setting is a domain transfer scenario. For the same gesture recognition task, the proposed architecture is supervised with only simulator gesture annotation and test on the real robot.
We conducted extensive experiments to evaluate the proposed UDA model, which is trained only with the supervision of gesture annotations generated from the simulator and then tested on the real robot.

The dataset we utilize is the Desk dataset\cite{madapana2019desk}, the peg transfer task in Desk dataset has the following variations that could lead to kinematic or visual domain shift between simulator and real robot: 1) Initial and final positions of objects on the pegboard, 2) direction of the transfer(objects on the left side are transferred to the right and vice versa) and 3) position/orientation of the pegboard(left/right) maneuvers. From the DESK dataset, we used Taurus simulator video sequences of S1-S5, Taurus real robot S1-S8. A five-fold cross-validation approach is used with a data split of $4:1$ for training and testing respectively. 
\begin{comment}
More importantly, the transfer process is conducted in an unsupervised manner where the surgeme annotation is only provided in simulator while not be provided in real robot. 
From the DESK dataset, we use Taurus simulator $S1-S5$ and Taurus $S1-S8$ ($S\text{x}$ indicates different pegboard transfer settings and the transfer task which can be referred to~\cite{madapana2019desk} for detailed dataset description). \\
\multicolumn{1}{c||}{\begin{tabular}[l]{@{}c@{}}Kin based\\ DA (ours) \end{tabular}}    & Kin                              & $68.57\pm0.05$          &23.59      &  $69.53\pm0.76$ & 20.44 &  $68.55\pm0.07$ & 23.84 &  $52.96\pm0.27$ & 26.65 &$68.56\pm0.34$ & 30.88  \\  
\hline
\end{comment}

To quantitatively analyze the performance of our method, we employ Accuracy (ACC), Precision (PR), Recall (RE), Jaccard (JA) and F1score (F1) to comprehensively validate the effectiveness of the proposed method. 
\begin{comment}
The PR, RE, JA and F1 Score are computed in phase-wise, and defined as:
\begin{equation}
\label{eq:eva}
\centering
\begin{gathered}
\mathrm{PR}=\frac{|\mathrm{GT} \cap \mathrm{P}|}{|\mathrm{P}|}, ~ \mathrm{RE}=\frac{|\mathrm{GT} \cap \mathrm{P}|}{|\mathrm{GT}|}, \\
\mathrm{JA}=\frac{|\mathrm{GT} \cap \mathrm{P}|}{|\mathrm{GT} \cup \mathrm{P}|}, ~
\mathrm{F1}=\frac{2}{\frac{1}{\mathrm{PR} } + \frac{1}{\mathrm{RE}}}, 
\end{gathered}
\small
\end{equation}
\end{comment}

\subsection{Comparisons with State-of-the-art Approaches}
We compare the proposed methods with state-of-the-art approaches as well as a baseline.
Table.~\ref{table:overallresults} shows the experimental results. 
%on the real robot using our KV-Sim2Real unsupervised domain adaptation (UDA) framework and the effectiveness of UDA.
The first two rows are the results without domain adaptation (DA), where the model is trained on simulator and tested on the real robot.
The Baseline model is based on our proposed deep learning framework in this paper but removes the domain adaptation task, which only conducts gesture recognition task. RF~\cite{rahman2019transferring} is one previous work and utilizes random forest for gesture recognition task.
Both two methods apply only kinematic data as input.
The rest methods all conduct UDA and can be categorized into three types: 1) visual (image / video) feature based DA (\cite{li2018adaptive}~\cite{saito2018maximum}); 2) kinematic (Kin) feature based DA (\cite{chen2019temporal}); 3) KV based DA (employ both visual and kinematic information).

\begin{comment}
Note that the proposed model is the first one to exploit both visual and kinematic information to transfer the knowledge learned from a simulator to a real robot.
First, the domain gap between simulator and real robot is very large even conduct the same task, with model trained on simulator applying on real robot can only get 44.30\% accuracy for surgeme recognition. It means that the kinematic or visual representation has discrepancy in two domain, so it is necessary to align the feature from different domain(domain adaption). 
\end{comment}

There are several key observations from the results.
First, the results show that the feature domain gap is larger in visual data than kinematic data but the methods utilize either visual domain or kinematic domain for UDA is not able to achieve satisfactory results. 
%
%More specifically, the image based DA method AdaBN~\cite{li2018adaptive} and MCD~\cite{saito2018maximum} conduct adaptation on image-level and video based DA method T3AN~\cite{chen2019temporal} conduct adaptation on video level. 
%
As shown in Table.~\ref{table:overallresults}, the results of~\cite{li2018adaptive}~\cite{saito2018maximum}~\cite{chen2019temporal} decrease from 44.30\% to 14.81\%, 16.03\% and 18.23\%, when compared to the Baseline.
In contrast, the MDO-K DA (motion direction oriented kinematic feature alignment, part C of Method section) promotes the ACC to 51.17\% by 6.19\% gain. 
The performance increase of applying kinematic DA and the decrease of applying visual DA indicates that the feature domain gap is larger in visual than kinematic data.
In this regard, transferring visual level feature is more difficult.
Second, the KV-Relation-ATT feature can further promote the adaptation performance. 
After adding KV-Relation-ATT adaptation on MDO-K adaptation, MDO-K + KV-Relation-ATT has a better performance compared than the MDO-K, which verifies that KV-Relation-ATT can let the model learning the correlation between kinematic and visual at the same time step $t$ and act as a stricter constraint for source to target feature domain adaptation compared with single MDO-K adaptation. That is, not only align the kinematic information but also align the co-occurrence of kinematic and visual information for learned features between domains.
\begin{comment}
combination of the kinematic as well as visual feature at same time would let the model not only focus on the co-occurrence feature between kinematic and visual feature.
\end{comment} This leads the model to learn a stronger kinematic motion but also focus on the visual motion, so the KV-relation is a new feature to represent the motion.
\begin{comment}
3) Aligning task-related feature is significant in DA. Our MDO-K DA method which align motion feature outperform `Baseline' training which only learning axis feature. It indicates that aligning task-related feature is essential in DA task.  
\end{comment}

\begin{comment}
\multicolumn{1}{c|}{\begin{tabular}[c]{@{}c@{}}Kin based\\ DA (ours) \end{tabular}}    & Kin DA                             & 68.57\pm0.05          &       \\ \hline
\end{comment}

\begin{figure}[ht]
	\centering
	{\includegraphics[width=1\linewidth]{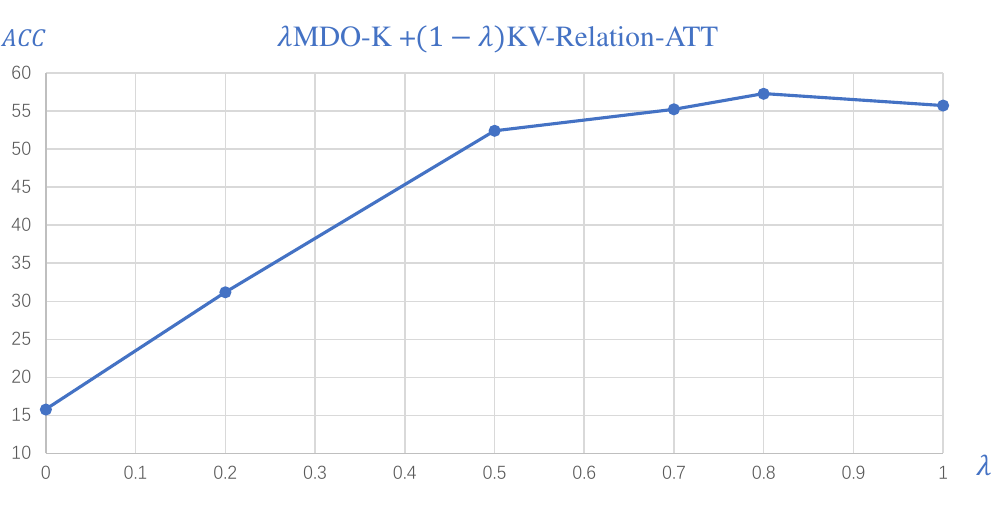}}
	\caption{Results of varying hyper-parameter $\lambda$ for balancing MDO-K and KV-Relation-ATT.}
	\label{Fig:lambda}
\end{figure}

\begin{figure*}[h!]
	\centering
	{\includegraphics[width=1\linewidth]{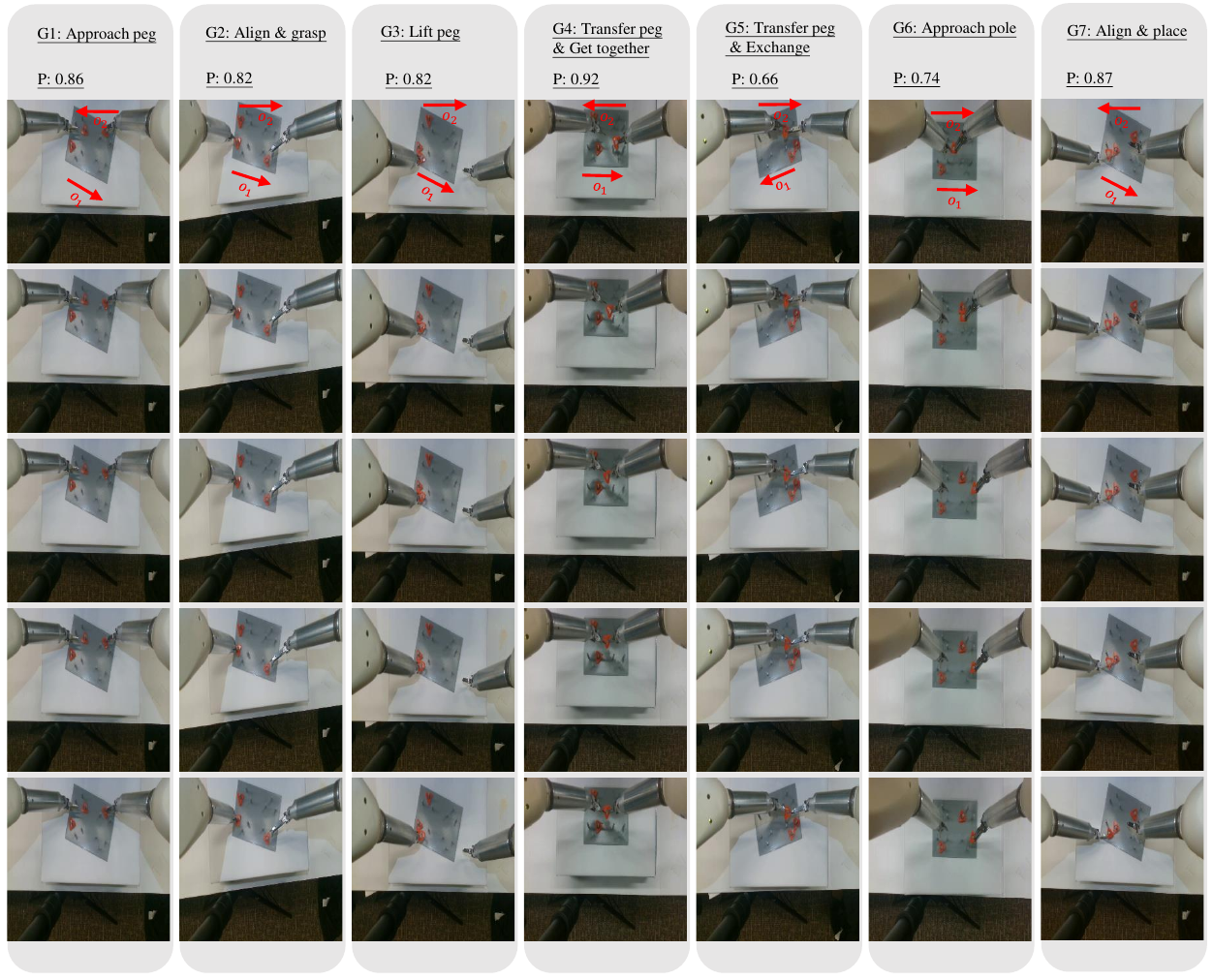}}
	\caption{Successful transfer cases on real robot with different domain gap situations. Each column is a sequential gesture process. $o_1$ represents the orientation of the pegboard, $o_2$ represents the orientation for peg transfer(from left to right or right to left). We can see that the successful case can handle different $o_1$ and $o_2$, which corresponds to variant domain gap situations with high model prediction probabilities (P).}
	\label{Fig:pro}
\end{figure*}
\begin{table}[]
	\centering
	\caption{Ablation study on the pattern of combing kinematic and visual DA. (mean$\pm$std., \%). '+' means multi-task learning results.}
	\begin{tabular}{l||l||l}
		& Accuracy         & Gain  \\ \hline 
		Baseline     & $44.98\pm2.00$ & -     \\ \hline
		MDO-K     & $51.17\pm0.10$ & 6.19     \\ \hline
		MDO-K + T3AN & $46.25\pm3.46$ & 1.27  \\ \hline
		MDO-K + KV-Relation-ATT  & $57.89\pm0.39$ & 12.91 \\ \hline
	\end{tabular}
	\label{table:KinVisablation}
\end{table}

\begin{table}[]
	\centering
	\caption{Ablation study on kinematic information. Results of utilizing position vector and unit direction vector training on simulator and test on real robot without DA. (mean$\pm$std., \%).}
	\begin{tabular}{l||l}
		& Accuracy           \\ \hline
		Kin (position vector) & $44.98\pm2.00$   \\\hline
		Kin (unit relative direction vector)  & $50.42\pm0.93$  \\ \hline
	\end{tabular}
	\label{table:kinuse}
\end{table}

\subsection{Ablation Study and Analysis}
Table.~\ref{table:KinVisablation} shows the ablation study of how to align different modality features. 
The Baseline is single kinematic modality without DA while MDO-K + T3AN and MDO-K + KV-Relation-ATT are Kin+Vis DA which align feature from multi-modality. 
The multi-modality feature domain adaptation conducts separately within each modality feature. 
There are several observations: 1) Adding visual domain adaptation separately can not promote the DA results much higher under the large visual modality domain gap, because MDO-K + T3AN increase by only 1.27\% compared with no adaptation Baseline and decrease 4.92\% compared with kin-based MDO-K adaptation. The reason comes from the large domain gap of the visual modality and leads to over-fitting problem to the source visual modality when adding visual adaptation separately with the kinematic adaptation.
2) KV-Relation-ATT can handle domain gap of visual modality and promote MDO-K  from 51.17\% to 57.89\% (MDO-K + KV-Relation-ATT). 
The observation indicates that when transferring both kinematic and visual information, kinematic should be the primary modality to transfer because of its relatively smaller domain gap compared to the visual gap. 
So our proposed KV-Relation-ATT adaptation which not individually adapts visual information can remedy the visual gap. 
Table.~\ref{table:kinuse} shows the results of utilizing position vector and unit director vector to train on simulator and test on real robot. 
We found that the unit relative direction vector would have better representation ability compared with the position vector because it let the model focus on the direction instead of the distance value of the kinematic data.

Fig.~\ref{Fig:lambda} shows the results of varying hyper-parameter $\lambda$ in multi-task based domain adaptation MDO-K + KV-Relation-ATT (Equ.~\ref{equ:CE}). 
We sample $0.2,0.5,0.7,0.8$ for comparison. 
We find that performance increase a lot from $0.2$ to $0.5$, which means that DA on kinematic feature would be better than KV-Relation-ATT feature ($\lambda$ represents the `MDO-K' feature contribution for gesture recognition). 
When $\lambda > 0.5$, the performance gradually increases and decreases, which means that KV-Relation-ATT act as an assist DA task for the MDO-K DA process, so a good balance of MDO-K + KV-Relation-ATT feature can promote DA results, in our setting, is $0.8$.
Fig.~\ref{Fig:pro} is the successful transfer cases on real robot with different domain gap situations. We can observe that under variant pegboard orientation($o_1$), peg transfer orientation($o_2$) or different gestures, the model can successfully generalize the feature to real robot recognition, which verifies its practical usage on real world scenarios.
\begin{comment}
Aligning multi-modality feature is not definitely better than aligning single-modality feature. Here, MDO-K + Video DA increase only 1.27\% ACC, the reason is also because the large domain gap of visual modality. 2)
\end{comment}

\section{CONCLUSIONS AND DISCUSSIONS}
For automated surgical gesture recognition in robot-assisted minimally invasive surgery, existing methods focus only on training and testing data on the real robot, but in reality, simulator and real robot exist domain gap. In this paper, we introduce a domain adaptation framework that can synchronize transfer multi-modality knowledge (kinematic data as well as visual data) from simulator to real robot to remedy the domain gap between them for peg transfer task. We propose two alignments for DA: an MDO-K alignment is proposed aiming at transferring kinematic motion direction from simulator to real robot. Besides, a KV-Relation-ATT feature alignment is proposed for transfer the co-occurrence signal feature of kinematic and visual modality. 
In summary, the performance show that aligns both kinematic and visual modality can promote the domain adaptation result to the best. Also, kinematic data is easier to align compared with visual data.

For model generalization ability, our proposed MDO-K can generalize to other tasks or scenarios and is not limited to peg transfer task or the specific video background, because we utilize unit direction vector instead of position vector which shares the same measurement for object motion trajectory across different tasks or dataset. Also, the joint utilization of kinematic data and visual data by our proposed KV-Relation-ATT can handle the over-fitting problem to visual modality. So the good generalization ability of KV-Relation-ATT can handle the variant kinematic gap and visual gap in different tasks or scenarios.

\bibliographystyle{IEEEtran}
\bibliography{ref}

\end{document}